%% file: ijcai23-main.tex
\documentclass{article}
\usepackage{ijcai22}

\usepackage{times}
\usepackage{soul}
\usepackage{url}
\usepackage[hidelinks]{hyperref}
\usepackage[utf8]{inputenc}
\usepackage[small]{caption}
\usepackage{graphicx}
\usepackage{amsmath,amssymb}
\usepackage{amsthm}
\usepackage{booktabs,tabularx,siunitx}
\usepackage{algorithm}
\usepackage{algorithmic}

\usepackage{multirow}
\usepackage{hyperref}
\usepackage{xcolor}
\usepackage{textcomp}  
\usepackage{scalerel}  
\def\pos{\scalerel*{\includegraphics{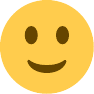}}{\textrm{\textbigcircle}}}
\def\neu{\scalerel*{\includegraphics{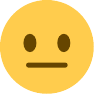}}{\textrm{\textbigcircle}}}
\def\neg{\scalerel*{\includegraphics{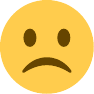}}{\textrm{\textbigcircle}}}

\usepackage{paralist}
\usepackage[shortlabels]{enumitem}

\DeclareMathOperator*{\argmax}{\arg\!\max}

\newcolumntype{C}{>{\centering\arraybackslash}X}

\newcounter{rowcntr}[table]
\renewcommand{\therowcntr}{\thetable.\arabic{rowcntr}}
\newcolumntype{N}{>{\refstepcounter{rowcntr}\therowcntr}c}
\AtBeginEnvironment{tabular}{\setcounter{rowcntr}{0}}

\title{The Emotions of the Crowd: Learning Image Sentiment from Tweets via Cross-modal Distillation}

\author{
Alessio Serra$^1$
\and
Fabio Carrara$^2$\and
Maurizio Tesconi$^3$\And
Fabrizio Falchi$^2$
\affiliations
$^1$Università di Pisa, Pisa, Italy\\
$^2$ISTI-CNR, Pisa, Italy\\
$^3$IIT-CNR, Pisa, Italy
}

\begin{document}

\maketitle

\begin{abstract}
Trends and opinion mining in social media increasingly focus on novel interactions involving visual media, like images and short videos, in addition to text.
In this work, we tackle the problem of visual sentiment analysis of social media images -- specifically, the prediction of image sentiment polarity.
While previous work relied on manually labeled training sets, we propose an automated approach for building sentiment polarity classifiers based on a cross-modal distillation paradigm;
starting from scraped multimodal (text + images) data,
we train a student model on the visual modality based on the outputs of a textual teacher model that analyses the sentiment of the corresponding textual modality.
We applied our method to randomly collected images crawled from Twitter over three months and produced, after automatic cleaning, a weakly-labeled dataset of $\sim$1.5 million images.
Despite exploiting noisy labeled samples, our training pipeline produces classifiers showing strong generalization capabilities and outperforming the current state of the art on five manually labeled benchmarks for image sentiment polarity prediction.
\end{abstract}

\section{Introduction}

Mining trends and opinions from social networks provides crucial information to help make strategic decisions in various fields.
Twitter data, for example, have been used to explain and predict social issues and user opinions on product brands and sales~\cite{jansen2009twitter,rui2013whose}, patient reactions to medicines~\cite{adrover2015identifying}, stock market movements~\cite{bollen2011twitter}, political performances and election outcomes~\cite{bermingham2010classifying,diakopoulos2010characterizing,mejova2013gop} and many others.
While most research in sentiment analysis from social-network data focused on text, online interactions increasingly involve visual media such as pictures, edited images, and short videos, putting more interest in visual sentiment analysis (VSA).
The main issue of state-of-the-art approaches for VSA is their strongly supervised nature:
manually labeling images for VSA is costly due to the subjectivity of image interpretation and the viewer's emotional response, thus requiring multiple labelers and limiting the dataset scale to a few thousand samples~\cite{katsurai2016image,you2016building}.
Moreover, natural distribution shifts occurring in opinions and trends would require repeating the labeling process periodically, which is unfeasible.

This paper proposes an automated approach to train models for visual sentiment analysis.
Specifically, we tackle the problem of predicting the average polarity of sentiments an image evokes to its viewers, usually coarsely estimated as being `positive', `neutral' or `negative'.
We propose an approach based on a cross-modal distillation method;
a pretrained textual sentiment predictor, acting as the teacher model, is distilled into a visual sentiment predictor using text-image pairs streamed from random-sampled multimodal posts as training samples.
The proposed approach is not fully unsupervised but rather based on distant supervision~\cite{mintz2009distant}, as we assume a pretrained textual teacher model that transfers knowledge to the student visual predictor.
However, the availability of self-supervised, easily fine-tunable language models 
makes it possible to harness the available resources for textual sentiment analysis and transfer their knowledge to the visual domain without additional labeling costs.
Moreover, our approach is employable in a continual learning setup, especially if employed with diachronic language models such as TimeLM~\cite{loureiro2022timelms}, providing an effective and cheap way to keep sentiment analysis tools up to date.

We apply our approach to random-sampled Twitter posts in three months (Apr-Jun 2022) and show that the obtained visual models outperform the current state of the art in five manually-annotated benchmarks for image sentiment polarity prediction.
We also contribute by releasing the code, trained models, and the set collected and preprocessed images ($\sim$1.5M) used in the experimental phase.

In summary, we contribute by
\begin{itemize}
    \item proposing a cross-modal distillation approach to train image sentiment polarity predictors without relying on manually labeled image datasets,
    \item testing the obtained models on five manually-labeled benchmarks and outperforming the current state of the art in five of them, and
    \item publicly releasing the code, the trained models, and the collected data ($\sim$3.7M images) used in our experiments.
\end{itemize}

\begin{figure*}
\centering
\includegraphics[width=\linewidth]{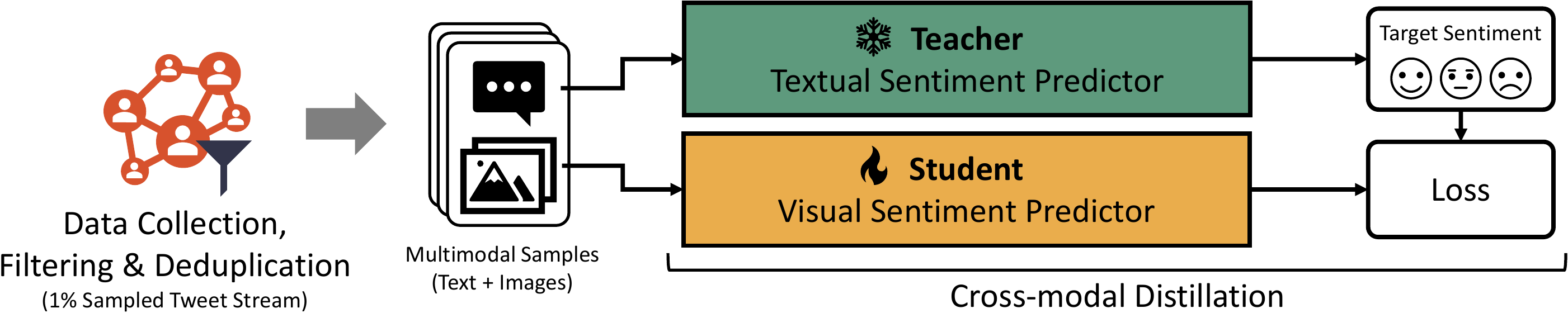}
\caption{Overview of the proposed approach. Tweets are filtered and deduplicated to keep multimodal samples with long-enough English text and at least one image. Then, we apply cross-modal distillation; given a text-image pair $(i,t)$, a visual student model is trained to predict from $i$ the same sentiment polarity of $t$ inferred by a textual teacher model (a pretrained textual sentiment classifier).
}
\label{fig:overview}
\end{figure*}

\section{Related Work}

Our main focus is purely visual sentiment analysis, where a judgment can be expressed by looking only at image pixels.
Other related tasks are also tackled, such as the well-explored textual-based sentiment analysis~\cite{liu2012survey,baccianella2010sentiwordnet,salur2020novel,khan2021exploiting} and directions also exploiting additional inputs or modalities~\cite{yu2021learning,hazarika2020misa,chauhan2019context,truong2019vistanet,li2018image,truong2023concept} or focusing on aspects different from sentiment, like virality or aesthetics~\cite{gelli2015image,khosla2014makes,totti2014impact}.

\paragraph{Visual Sentiment Analysis (VSA)}
Seminal approaches to visual sentiment analysis, mainly from 2010, were based on extracting handcrafted low-level features from input images based on color, texture, composition, and content characteristics.
For example~\cite{li2012scaring} merged SIFT descriptor, Gabor texture, and HSV color histogram to obtain a global feature vector and \cite{machajdik2010affective} extracted color, texture and harmonious composition from images.
Subsequent approaches leveraged mid-level features of images, such as the one proposed by \cite{borth2013large};
they built a visual sentiment ontology consisting of 3'000 adjective-noun pairs that express strong sentiment values and are related to an emotion, represented using the well-known ``Plutchik’s Wheel of Emotions" psychological model~\cite{robert1980emotion}.
The adjective-noun pair were used as keywords to get images from Flickr, which were then leveraged to train an individual tracker for each member of the ontology.
Subsequently, only reasonably performing detectors were selected to compose SentiBank, their proposed framework capable of extracting mid-level characteristics from images, which can be used as input for a sentiment classifier.
However, research has recently been geared towards deep learning models, which can automatically learn how to extract high-level visual characteristics from raw input data.
Most methods in this category rely on supervised transfer learning, exploiting various convolutional models like GoogleNet~\cite{islam2016visual}, AlexNet-inspired~\cite{campos2017pixels}, or custom architectures~\cite{you2015robust}.
One of the most recent approaches is \cite{wu2020visual}, which combines global and local features of the image using both a CNN and a saliency detector; in particular, salient sub-images are detected, and then an optimized VGGNet makes a prediction on both the entire image and the sub-images. Finally, the predictions are combined by a weighted sum to detect a positive, neutral, or negative sentiment polarity.

\paragraph{Dataset for VSA}
Even for VSA, models are often as good as their training data are.
The most used approach to build a dataset for a VSA task relies on manual annotation since it allows getting reliable, strong labels.
However, it is also costly due to the subjectivity of the sentiment we attach to samples, thus requiring more than one annotator to incorporate multiple perspectives into the labeling.
For this scope, many esearchers~\cite{you2015robust,borth2013large,you2016building,peng2015mixed,katsurai2016image} relied on crowdsourcing services, i.e., Amazon Mechanical Turk (AMT), to involve multiple labelers and ensure strong labels.
In addition, \cite{you2016building,peng2015mixed} select labelers based on their ability to classify feelings using a qualification test, ensuring cleaner labels.
However, scaling datasets beyond the order of tens of thousands of samples still requires a non-negligible effort.

\paragraph{Weak supervision}
Adopting weak supervision allows us to obtain much larger datasets at the cost of lowering the labeling quality and introducing label noise.
In the visual domain, this technique recently gained more and more attention.
For example, \cite{sun2017revisiting} exploited a complex mixture of raw web signals, connections between web pages, and user feedback to generate a huge image classification dataset, and \cite{mahajan2018exploring} relied on hashtag prediction on social media images.
For VSA, there are just a few examples. The approach of \cite{siersdorfer2010analyzing} assigns weak sentiment labels to images coming from Flickr based on image tags. Still, it is susceptible to noisy or missing tags and is biased by the tags' choice; similarly, \cite{2017_ICCV} assigns weak labels by analyzing the text content of tweets.
Our approach follows this direction by crawling randomly sampled multimodal data from social media streams, but the supervision signal is obtained by distilling a textual sentiment predictor into a visual model.

\section{Methodology}
As done in previous work, we formulate image sentiment polarity prediction as an $N$-way image classification problem.
Our objective is to learn an image classifier that assigns the correct sentiment label out of $N$ possible labels to an input image without resorting to supervised training and, thus, an expensive manual annotation of images.
To do so, we propose an automatic approach organized in two steps; a) \textit{data collection, filtering, and deduplication} and b) \textit{cross-modal distillation}.
Figure~\ref{fig:overview} schematizes our proposal.
We further describe each step in the following subsections.

\subsection{Data Collection, Filtering, and Deduplication}
This first step aims to construct a data stream to fuel the subsequent learning step.
We crawl data from a social network of interest by collecting random posts in a specified period.
In this work, we demonstrate our proposal on Twitter, but in principle, any platform providing access (free or paid) to large volumes of randomly-sampled posts can be used.

To subsequently apply a cross-modal paradigm, we are interested in filtering out samples having only a single modality in favor of ones containing both text and one or more images.
We apply the same filtering steps applied in \cite{2017_ICCV} and keep only tweets that
\begin{inparaenum}[a)]
\item have a text comprised of 5 or more words in the English language,
\item have at least one image, and
\item are not retweets.
\end{inparaenum}
We thus obtain a set $S = \{s_j\}_{j=1}^M$ of text-image pairs $s_j = (t_j, i_j)\,, t_j \in T\,, i_j \in I$, where $T$ and $I$ respectively indicate the space of texts and images. We indicate with $M$ the number of samples at the end of the collection campaign, but in an online learning configuration, $S$ constitutes an infinite data stream.

Due to the virality of some contents, a non-negligible part of posts and corresponding images crawled end up duplicates or near-duplicate images.
To make the process leaner and obtain a more varied stream of visual data, we drop samples having the same or nearly-same content in the visual medium.
Specifically, we assume two samples $s_1 = (t_1, i_1)$ and $s_2 = (t_2, i_2)$ are duplicates if $\cos(\Phi(i_1), \Phi(i_2)) > \tau$, where $\Phi(i) \in \mathbb{R}^n$ is a feature vector extracted from the image $i$ by a general-purpose pretrained visual model $\Phi$, and $\tau$ is an empirically-chosen threshold.

\subsection{Cross-modal Distillation}

We set up a cross-modal student-teacher learning paradigm fed by data streaming from the previous step.

Let $g: T \to [0,1]^N$ a pretrained textual sentiment polarity predictor that maps an input text into an $N$-dimensional categorical distribution and similarly,
$f: I \to [0,1]^N$ an image classifier sharing the same label space as $g$.
Given a set of multimodal samples $S = \{s_j\}_{j=1}^M$
, we train the student model $f$ to align its prediction on the visual modality to the ones of the teacher model $g$ on the textual modality.
Formally, for a single text-images pair $s = (t, i)$, we minimize the following cross-entropy loss 
\begin{equation}
    \mathcal{L}(t,i) = \lambda(g(t)) \sum_{k=1}^N g_k(t) \log (f_k(i)) \,,
\end{equation}
where $g_k(t)$ and $f_k(i)$ indicate the $k$-th output of the teacher and student model, respectively, and
\begin{equation}\label{eq:filter}
    \lambda(g(t)) =
    \begin{cases}
        1 & \text{if } g_{\bar{k}}(t) \geq c_{\bar{k}}\,,\; \bar{k} = \underset{k}{\argmax}\ g_k(t)\\
        0 & \text{otherwise }
    \end{cases}
\end{equation}
is a multiplier that filters out low-confidence samples, as it sets the sample loss to zero if the probability of the most confident class $g_{\bar{k}}(t)$ is below a predefined threshold $c_{\bar{k}}$ that is defined for each possible class $\{c_j\}_{j=1}^N\,, c_j \in [0,1]$.
We define $\lambda(g(t))$ to represent a generic weighting scheme for training samples.
Equation~\ref{eq:filter} represents a hard gating strategy based on the teacher’s confidence.
In future work, we plan to explore other formulations, such as soft gating.
During training, the teacher model $g$ is frozen, and only $f$ is updated by gradient-based optimization until convergence.

\section{Experiments}

\subsection{Experimental Setup}

\paragraph{Data Collection}
We collected roughly 3M tweets with 3.7M images (1.26 images per tweet on average) in three months between April and June 2022.
Crawling was implemented via the Twitter API Volume Streams\footnote{\url{https://developer.twitter.com/en/docs/twitter-api/tweets/volume-streams/introduction}} that provides a streaming endpoint delivering roughly a 1\% random sample of the global and publicly available tweets in real-time.
For deduplication, we choose an ImageNet-pretrained ResNet-50 as feature vector extractor $\Phi$;
specifically, we use the max-pooled output of the sixth residual block as feature vector and mark tweets as duplicates if their image contents have very-high cosine similarity ($\tau = 0.98875$).
Deduplication yielded a $\sim$22\% reduction of the image set, which went from  $\sim$3.7M to $\sim$2.9M.
In Table~\ref{tab:dataset_composition}, we report a summary of collected data broken down by the three sentiment polarity classes induced by the teacher model chosen in our experimentation (more on this in the following subsections).
As an additional source of samples, we also employ B-T4SA~\cite{2017_ICCV} --- a set of 470586 text+images tweets collected following the same crawling rules between July and December 2016.
Following a chronological order, we refer to the B-T4SA dataset as \textsf{A} and our newly collected dataset as \textsf{B}.

\paragraph{Teacher Architecture}
Among many approaches proposed in the literature for textual sentiment analysis, for the teacher model, we choose a model from Time-LMs~\cite{loureiro2022timelms} --- a family of models trained with a \textit{continual learning} approach.
It comprises a BERT-based model trained on real-time Twitter data and periodically released, enabling diachronic specialization that is particularly relevant in the social media domain where the topic of discussion changes rapidly, as well as slang and language used.
For instance, a model trained before 2019 would not be aware of the meaning of neologisms such as \textit{“COVID-19”} or the different feelings related to \textit{“swabs”} or \textit{“variant”} that we give after the pandemic.
We select the Time-LM model released at the end of June 2022, fine-tuned for sentiment analysis on the TweetEval benchmark~\cite{barbieri2020tweeteval} available in the TweetNLP library~\cite{camacho-collados-etal-2022-tweetnlp} 
This choice also sets the granularity of the prediction ($N=3$), as the model has three possible outputs; `positive', `neutral', or `negative' sentiment polarity.

\paragraph{Student Architecture}
As the visual student model, we select a Vision Transformer (ViT)~\cite{dosovitskiy2020image} with the final head adjusted to output $N=3$ logits.
We start training from the publicly available checkpoints pretrained on \textit{Imagenet-21k} and on \textit{Imagenet-1k}.
During training, we employ data augmentation on the visual pipeline by applying random horizontal flips, shifts, and rotations.
Optimization is carried out using the Adam optimizer with an initial learning rate of $10^{-4}$, $\epsilon=10^{-7}$, $\beta_1=0.9$, and $\beta_2=0.999$.

\begin{table}
\caption{Summary on collected and preprocessed training data broken down by the sentiment polarity assigned by the teacher model (TimeLM Jun-2022 fine-tuned on TweetEval).}
\label{tab:dataset_composition}
\newcommand{\M}[2]{\multicolumn{#1}{c}{#2}}
\begin{tabularx}{\linewidth}{X*3{S[table-format=7.0]}}
\toprule
                   & \M{2}{\textbf{Collected}}            & \M{1}{\textbf{Deduplicated}} \\
                     \cmidrule(lr){2-3}                     \cmidrule(lr){4-4}
\textbf{Sentiment} &  \M{1}{\# tweets} & \M{1}{\# images} & \M{1}{\# images} \\
\midrule
Positive           &           1206158 &          1593484 &  1299916 \\
Neutral            &           1403683 &          1708195 &  1293259 \\
Negative           &            356002 &           433172 &   329395 \\ \midrule
Total              &           2965843 &          3734849 &  2922568 \\
\bottomrule
\end{tabularx}
\end{table}

\subsection{Benchmarks}
\label{sec:benchmark}
To test the effectiveness of the proposed cross-modal training process, we evaluate our models on the benchmarks for image sentiment polarity prediction manually annotated via Amazon Mechanical Turk (AMT).
We consider
\begin{inparaenum}[a)]
    \item Twitter Dataset (TD)~\cite{you2015robust},
    \item Flickr\&Instagram (FI)~\cite{you2016building}, and
    \item EmotionROI~\cite{peng2015mixed}.
\end{inparaenum}
TD provides three benchmarks corresponding to three different levels of label agreement, i.e.,  where at least five, four, or three AMT workers agreed on the labels assigned to images.
The other datasets provide a single set of images with already aggregated labels.
TD provides binary labels (`positive' or `negative') for sentiment polarity.
Thus we mask the neutral class output of our models and take the maximum confidence among positive and negative outputs.

\begin{figure*}
\newcommand{\pic}[2]{\fcolorbox{#1}{white}{\includegraphics[height=2.9cm]{images/samples/#2.jpg}}}
\centering
\pic{green}{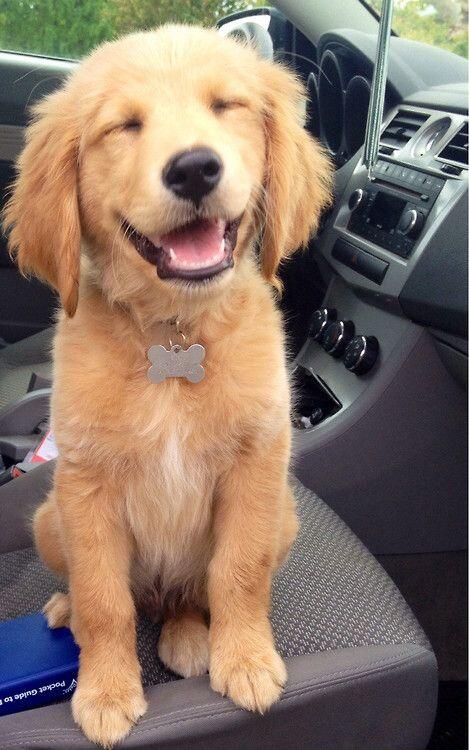}
\pic{red}{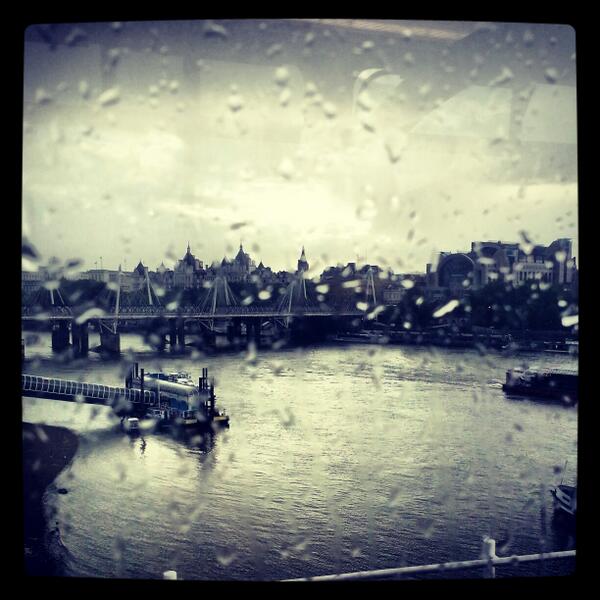}
\pic{green}{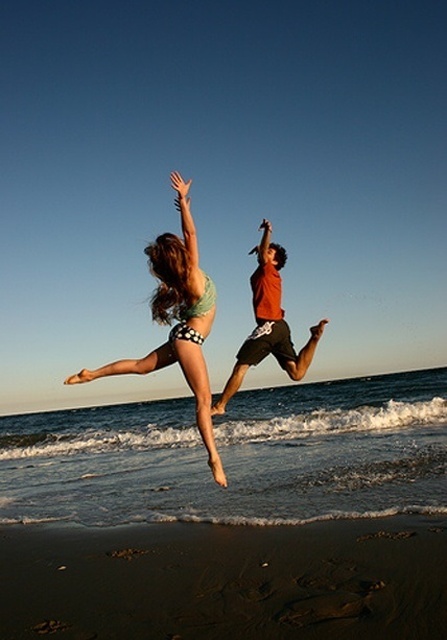}
\pic{red}{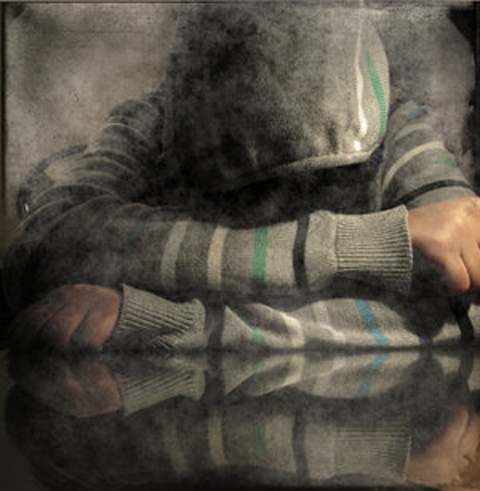}
\pic{green}{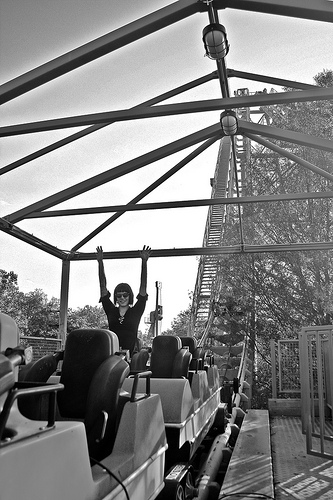}
\pic{red}{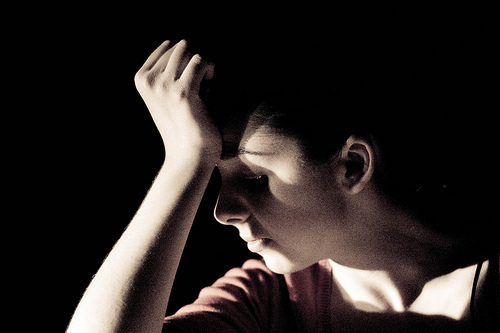}
\caption{Samples from the manually-annotated benchmark used for evaluation. From left to right, we show a \fcolorbox{green}{white}{positive} and a \fcolorbox{red}{white}{negative} sample for TD, EmotionROI, and FI benchmarks.}
\label{fig:samples}
\end{figure*}

FI and EmotionRoI provide fine-grained sentiment annotations and are used in literature as sentiment polarity benchmarks by mapping labels into two `positive' and `negative' polarities~\cite{wu2020visual}.
In particular, for dataset FI, the emotions of \textit{Awe}, \textit{Amusement}, \textit{Excitement}, and \textit{Contentment} are mapped to the `positive' polarity while \textit{Fear}, \textit{Disgust}, \textit{Sadness}, and \textit{Anger} to `negative'.
For EmotionROI, \textit{Anger}, \textit{Disgust}, \textit{Fear}, and \textit{Sadness} are relabeled as `negative', and \textit{Joy} and \textit{Surprise} as `positive'.
Table~\ref{tab:manual_dataset} reports the characteristics of each dataset, and Figure~\ref{fig:samples} shows some examples.
We adopt TD for preliminary experiments and ablation studies while we compare the best-performing models with other state-of-the-art methods on all the mentioned benchmarks.

\begin{table*}
\caption{Summary of manually annotated benchmarks used for evaluation. 
We report the number of samples remapped into `positive' and `negative' polarity for datasets with more than two classes.}
\label{tab:manual_dataset}
\centering
\renewcommand{\b}{\bfseries}
\newcommand{\M}[2]{\multicolumn{#1}{c}{#2}}
\begin{tabular}{lcc*3r}
\toprule
                                         &               &                   & \M{3}{\b \# Images}  \\ \cmidrule(l){4-6}
\b Dataset                               & \b \# Classes & \b \# AMT Workers & \pos & \neg & Tot.   \\ \midrule
Twitter Dataset \cite{you2015robust}   & 2             & 5                 & 769   & 500  & 1,269  \\
EmotionROI \cite{peng2015mixed}          & 6             & 432               & 660   & 1,320 & 1,980  \\
Flickr\&Instagram \cite{you2016building} & 8             & 1,000              & 16,430 & 6,878 & 23,308 \\
\bottomrule
\end{tabular}
\end{table*}

\subsection{Ablation study}
In this section, we evaluate how aspects such as data freshness, data filtering, and model architecture can affect the effectiveness of trained models.
We perform experiments varying the inputs and hyperparameters of our approach and producing several models.
We apply the obtained models on the TD benchmark in a zero-shot configuration (no learning on benchmark data is performed) and measure the classification accuracy.
Table~\ref{tab:ablation-super} reports all the obtained results we discuss below.

\begin{table*}
\caption{Ablation study. Accuracy on the three Twitter Dataset benchmarks (at-least-five-, four-, and three-agreement subsets). %
\textsf{A} = Set of tweets collected in Jul-Dec 2016 by~\protect\cite{2017_ICCV}. %
\textsf{B} = Set of tweets collected in Apr-Jun 2022 by us. %
Confidence filtering columns report the values used for the parameters $\{c_j\}_{j=1}^3$. %
}
\label{tab:ablation-super}
\centering
\input{tab-ttd-results.tex}
\end{table*}

\paragraph{Confidence Filtering}
In this experiment, we fix the input set (\textsf{A}) and the student model architecture (ViT Base with 86M parameters and a patch size of 32) and run our pipeline with or without confidence filtering, i.e., setting $c_j = .70\,, \forall j$ or $c_j = 0\,, \forall j$ in Equation~\ref{eq:filter}.
Comparing rows \ref{exp:A} and \ref{exp:Ac} in Table~\ref{tab:ablation-super}, we note that masking low-confidence samples in the student loss helps increase accuracy by 1--2\%.

\paragraph{Input Data}
In experiment~\ref{exp:Bf}, we repeat experiment~\ref{exp:Ac}, swapping the set \textsf{A} collected in 2016 with the one collected by us in 2022 (\textsf{B}).
We observed a small accuracy loss in the five-agree benchmark. Despite having more images, our set is more unbalanced towards positive and neutral classes with respect to \textsf{A}, which the original authors already balanced during data cleaning.
Indeed, setting higher confidence thresholds for those classes (experiment~\ref{exp:B}) mitigates this problem and provides additional improvements also to the lower-agree benchmarks.
Combining the two sets (experiment~\ref{exp:AB}) further increases performance by $\sim$2\%.

\paragraph{Student Architecture}
We evaluate scaling the model parameters and the patch size of the student ViT architecture.
Starting with the configuration of experiment~\ref{exp:AB}, in experiment~\ref{exp:ABL}, we swap the student model for the larger ViT-Large (307M parameters, $\sim$3.5x more than ViT-Base), while in experiments~\ref{exp:AB16} and \ref{exp:ABL16}, we repeat experiments~\ref{exp:AB} and \ref{exp:ABL} decreasing the input patch size from 32 to 16 (4x larger input sequences).
Decreasing patch size alone (\ref{exp:AB16}) is more effective than increasing model parameters (\ref{exp:ABL}), as the visual model can grasp finer details of the input image.
Scaling both dimensions together (\ref{exp:ABL16}) produces our best-performing configuration, confirming recent findings~\cite{kaplan2020scaling}.

\begin{figure*}
\centering
\includegraphics[height=3.07cm]{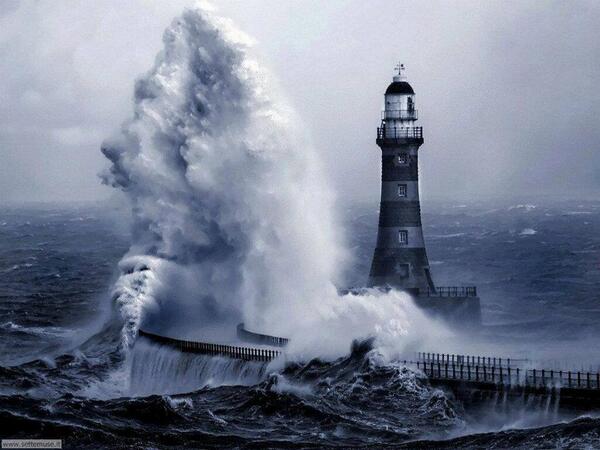}\hfill%
\includegraphics[height=3.07cm]{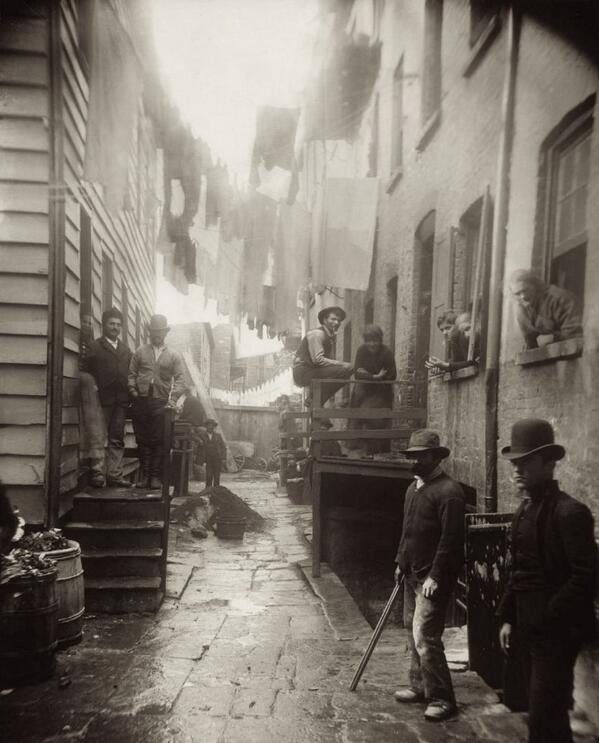}\hfill%
\includegraphics[height=3.07cm]{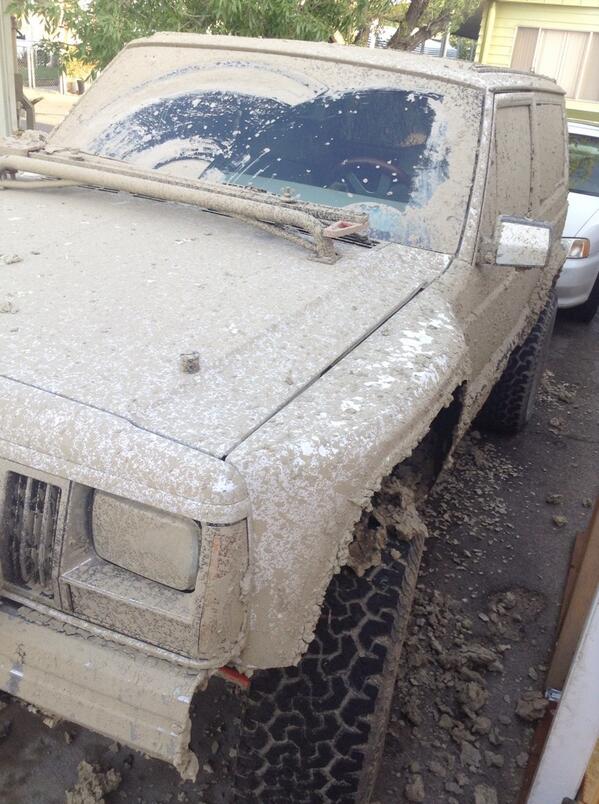}\hfill%
\includegraphics[height=3.07cm]{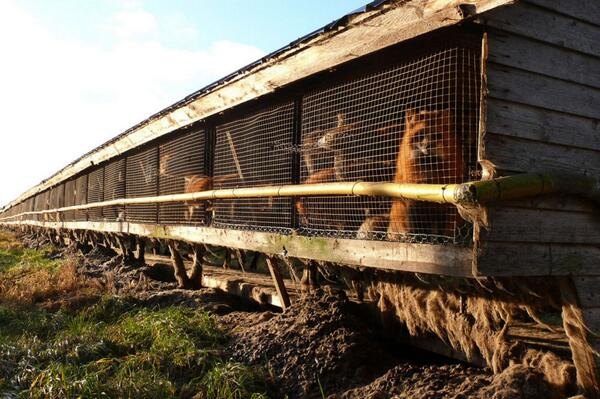}\hfill%
\includegraphics[height=3.07cm]{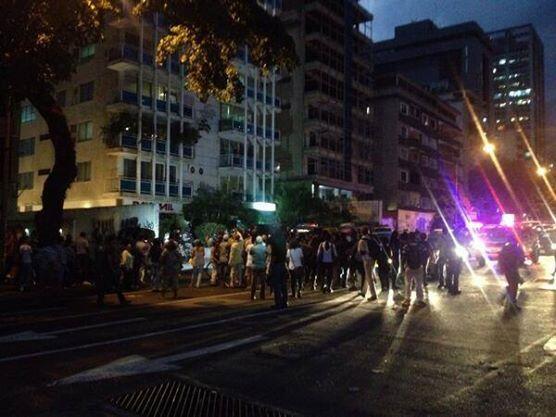}\\[.25ex]%
\includegraphics[height=3.73cm]{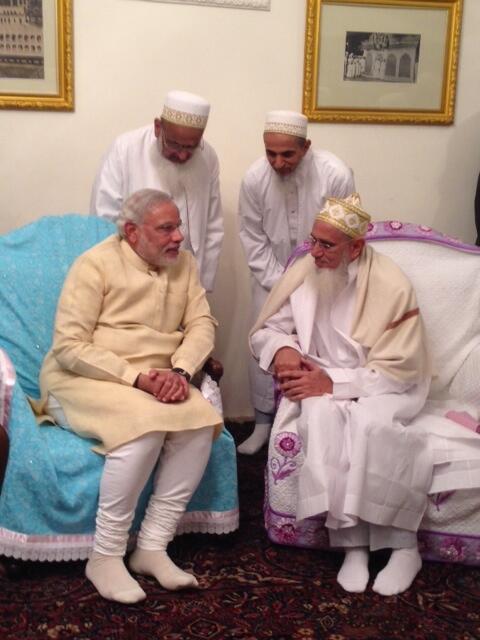}\hfill%
\includegraphics[height=3.73cm]{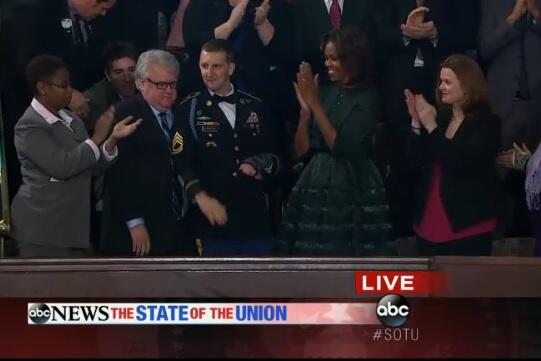}\hfill%
\includegraphics[height=3.73cm]{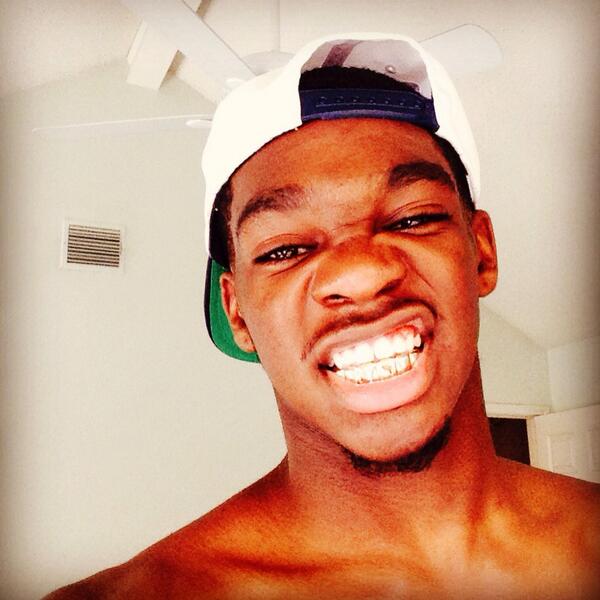}\hfill%
\includegraphics[height=3.73cm]{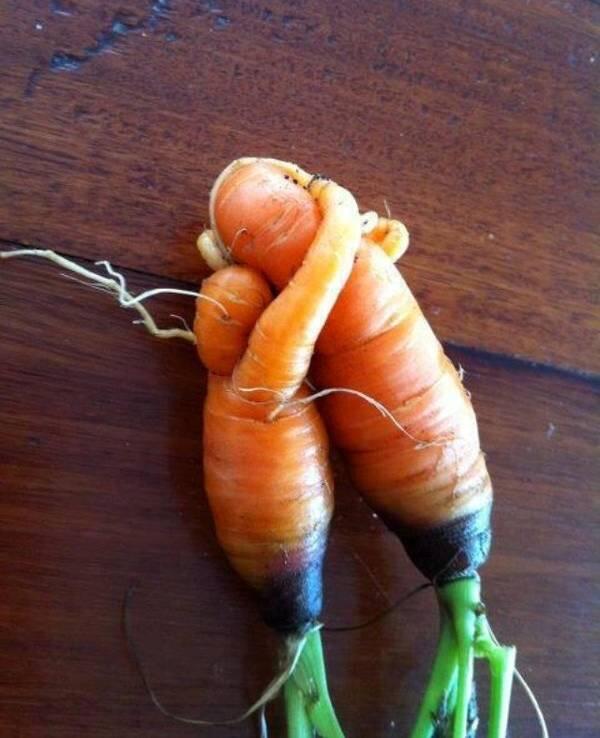}\hfill%
\includegraphics[height=3.73cm]{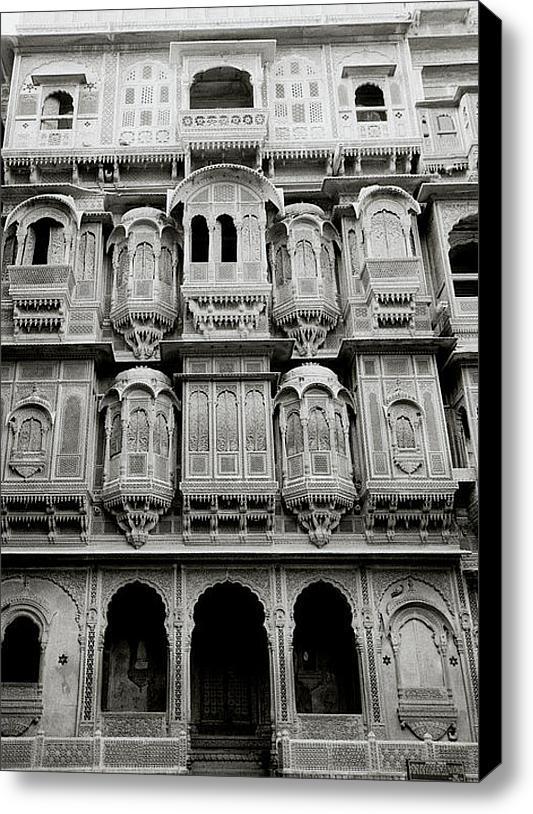}%
\caption{
Some cherry-picked examples of failure cases of our best model (ViT-L/16) on the Twitter Dataset benchmark.
The first row contains negative-labeled samples misclassified as positives, and the second row contains positive-labeled ones misclassified as negatives.
Note that several samples appear ambiguous even to a human labeler due to personal sensibility.
}
\label{fig:failure}
\end{figure*}

\subsection{Comparison with State of the Art}

\begin{table*}
\caption{Accuracy on standard benchmarks compared with state-of-the-art image sentiment polarity predictors.}
\label{tab:sota}
\input{tab-sota-results.tex}
\end{table*}

We compare our best model (ViT-L/16 trained on \textsf{A+B}) to state-of-the-art methods on the five manually-labeled benchmarks for image sentiment polarity described in Section~\ref{sec:benchmark}.
For a fair comparison, we follow the evaluation protocol of previous work~\cite{wu2020visual} that includes fine-tuning the models on the benchmark data.
Specifically, for TD and Emotion ROI, 5-fold cross-validation is performed, while for FI, models are trained on five random splits with 80/5/15 proportions of training/validation/test subsets.
For each benchmark, we measure the mean and standard deviation of the accuracy on the test splits.

As seen in Table~\ref{tab:sota}, our models outperform or are comparable to other state-of-the-art methods in all benchmarks.
Without fine-tuning, our models still obtain satisfactory results.
For the TD benchmark, which shares a data distribution similar to the one of the crawled data used, our model achieves an accuracy comparable to fine-tuned state-of-the-art models, even outperforming them on the 3-agreement subset.
On the other hand, the distribution shift between Twitter images and the Emotion ROI and FI benchmarks are too significant to ensure generalization.
We deem the culprit to be the class distribution for Emotion ROI, which privileges a negative sentiment polarity contrarily to other datasets, and the domain gap for FI, where images comprise more high-quality artistic pictures rather than synthetic/edited images and pictures taken with a smartphone.
However, fine-tuning reduces these gaps, showing that the knowledge in our model can be easily transferred to other domains.

In Figure~\ref{fig:failure}, we report some cherry-picked failure cases of our best model (non-finetuned ViT-L/16) on the Twitter Dataset benchmark.
Most failure cases comprise very subjective samples, for which the correct label is not immediately clear, even for a human judge.

\section{Conclusion}

We presented an automated approach to obtain trained models for visual sentiment analysis targeted for social media mining.
Harnessing existing resources for textual sentiment analysis, the proposed cross-modal distillation approach can produce robust models for image sentiment polarity prediction without any human intervention in data collection or labeling.
The experimental phase on Twitter data showed that our models reached a significant performance on manually-annotated benchmarks, setting the new state of the art on five of them.
All the collected data, the annotated datasets, and the trained models will be publicly available.
Moreover, the presented pipeline enables the production of visual diachronic models via continual learning from streaming social media data.

However, several limitations remain to be tackled.
One of the main issues (and thus motivation for future work) is the lack of zero-shot generalization to other domains, i.e., social media.
Although finetuning our models demonstrated a great transferability of the knowledge extracted from Twitter data, applying the model as-is yielded a satisfactory performance only on same-domain data.
Drawing data from a stream of multiple social media would improve zero-shot generalization and enable experimentation on larger scales.
Moreover, confidence filtering is still manually tuned for the particular distribution of input data, while an adaptive online balancing of samples will be explored in future work.

\section*{Ethical Statement}
The use of sentiment analysis by large corporations to achieve commercial benefit poses an ethical issue, as it runs the risk of causing detrimental effects on individuals or groups of people.
Moreover, the proposed method is intended to be used in conjunction with ethical web scraping.
The experiments reported in this work have been conducted exploiting the Twitter developer API complying with their Terms of Service.

\section*{Acknowledgments}

This work was partially funded by:
AI4Media - A European Excellence Centre for Media, Society and Democracy (EC, H2020 n. 951911);
SERICS (PE00000014) under the MUR National Recovery and Resilience Plan funded by European Union - NextGenerationEU.

\bibliographystyle{named}
\bibliography{biblio}

\end{document}

%% file: tab-ttd-results.tex

\renewcommand{\b}{\bfseries}
\newcommand{\M}[2]{\multicolumn{#1}{c}{#2}}
\begin{tabularx}{.75\linewidth}{Nc*3Cc*3c}

\toprule
 \M{1}{}  && \M{3}{\b Confidence Filter} && \M{3}{\b Twitter Dataset} \\
        \cmidrule(lr){3-5}                 \cmidrule(lr){7-9}
\M{1}{\b \#}      & \b Dataset   & \pos & \neu & \neg & \b Student Model & 5 agree & $\ge$4 agree & $\ge$3 agree \\ \midrule
\label{exp:A}     & \textsf{A}   & -    & -    & -    & B/32 & 82.2 & 78.0 & 75.5 \\
\label{exp:Ac}    & \textsf{A}   & .70  & .70  & .70  & B/32 & 84.7 & 79.7 & 76.6 \\ \midrule
\label{exp:Bf}    & \textsf{B}   & .70  & .70  & .70  & B/32 & 82.3 & 78.7 & 75.3 \\
\label{exp:B}     & \textsf{B}   & .90  & .90  & .70  & B/32 & 84.4 & 80.3 & 77.1 \\
\label{exp:AB}    & \textsf{A+B} & .90  & .90  & .70  & B/32 & 86.5 & 82.6 & 78.9 \\ \midrule
\label{exp:ABL}   & \textsf{A+B} & .90  & .90  & .70  & L/32 & 85.0 & 82.4 & 79.4 \\
\label{exp:AB16}  & \textsf{A+B} & .90  & .90  & .70  & B/16 & 87.0 & 83.1 & 79.4 \\
\label{exp:ABL16} & \textsf{A+B} & .90  & .90  & .70  & L/16 & 87.8 & 84.8 & 81.9 \\
\bottomrule
\end{tabularx}\\

%% file: tab-sota-results.tex
\renewcommand{\b}{\bfseries}
\newcommand{\M}[2]{\multicolumn{#1}{c}{#2}}

\centering
\begin{tabularx}{.8\linewidth}{X*5c}
\toprule
               & \M{3}{\textbf{Twitter Dataset}} \\
                 \cmidrule(lr){2-4}                                   
\b Model       & 5 agree & $\geq$4 agree & $\geq$3 agree & \b Emotion ROI & \b FI\\ 
\midrule
\cite{chen2014deepsentibank}* & 76.4 & 70.2 & 71.3 & 70.1 & 61.5 \\
\cite{you2015robust}*         & 82.5 & 76.5 & 76.4 & 73.6 & 75.3 \\
\cite{jou2015visual}$^\dagger$& 83.9$\pm$0.3 \\
\cite{2017_ICCV}              & 89.6 & 86.6 & 82.0 \\ 
\cite{yang2018visual}*        & 88.7 & 85.1 & 81.1 & 81.3    & 86.4 \\
\cite{wu2020visual}           & 89.5 & 87.0 & 81.7 & \b 83.0 & 88.8 \\ \midrule
ViT-L/16 (no fine-tuning)     & 87.8 & 84.8 & 81.9 & 64.1    & 76.0 \\
ViT-L/16        & \b 92.4$\pm$2.0 & \b 90.2$\pm$2.0 & \b 86.3$\pm$3.0  & \b 83.9$\pm$1.0 & \b89.4$\pm$0.1 \\
\midrule
\multicolumn{6}{l}{*As reported by~\cite{wu2020visual}. $^\dagger$As reported by~\cite{campos2017pixels}.}
\end{tabularx}

%% file: ijcai23-main.bbl
\begin{thebibliography}{}

\bibitem[\protect\citeauthoryear{Adrover \bgroup \em et al.\egroup
  }{2015}]{adrover2015identifying}
Cosme Adrover, Todd Bodnar, Zhuojie Huang, Amalio Telenti, Marcel Salath{\'e},
  et~al.
\newblock Identifying adverse effects of hiv drug treatment and associated
  sentiments using twitter.
\newblock {\em JMIR public health and surveillance}, 1(2):e4488, 2015.

\bibitem[\protect\citeauthoryear{Baccianella \bgroup \em et al.\egroup
  }{2010}]{baccianella2010sentiwordnet}
Stefano Baccianella, Andrea Esuli, Fabrizio Sebastiani, et~al.
\newblock Sentiwordnet 3.0: an enhanced lexical resource for sentiment analysis
  and opinion mining.
\newblock In {\em Lrec}, volume~10, pages 2200--2204, 2010.

\bibitem[\protect\citeauthoryear{Barbieri \bgroup \em et al.\egroup
  }{2020}]{barbieri2020tweeteval}
Francesco Barbieri, Jose Camacho-Collados, Leonardo Neves, and Luis
  Espinosa-Anke.
\newblock Tweeteval: Unified benchmark and comparative evaluation for tweet
  classification.
\newblock {\em arXiv preprint arXiv:2010.12421}, 2020.

\bibitem[\protect\citeauthoryear{Bermingham and
  Smeaton}{2010}]{bermingham2010classifying}
Adam Bermingham and Alan~F Smeaton.
\newblock Classifying sentiment in microblogs: is brevity an advantage?
\newblock In {\em Proceedings of the 19th ACM international conference on
  Information and knowledge management}, pages 1833--1836, 2010.

\bibitem[\protect\citeauthoryear{Bollen \bgroup \em et al.\egroup
  }{2011}]{bollen2011twitter}
Johan Bollen, Huina Mao, and Xiaojun Zeng.
\newblock Twitter mood predicts the stock market.
\newblock {\em Journal of computational science}, 2(1):1--8, 2011.

\bibitem[\protect\citeauthoryear{Borth \bgroup \em et al.\egroup
  }{2013}]{borth2013large}
Damian Borth, Rongrong Ji, Tao Chen, Thomas Breuel, and Shih-Fu Chang.
\newblock Large-scale visual sentiment ontology and detectors using adjective
  noun pairs.
\newblock In {\em Proceedings of the 21st ACM international conference on
  Multimedia}, pages 223--232, 2013.

\bibitem[\protect\citeauthoryear{Camacho-Collados \bgroup \em et al.\egroup
  }{2022}]{camacho-collados-etal-2022-tweetnlp}
Jose Camacho-Collados, Kiamehr Rezaee, Talayeh Riahi, Asahi Ushio, Daniel
  Loureiro, Dimosthenis Antypas, Joanne Boisson, Luis Espinosa-Anke, Fangyu
  Liu, Eugenio Mart{\'\i}nez-C{\'a}mara, et~al.
\newblock {T}weet{NLP}: {C}utting-{E}dge {N}atural {L}anguage {P}rocessing for
  {S}ocial {M}edia.
\newblock In {\em Proceedings of the 2022 Conference on Empirical Methods in
  Natural Language Processing: System Demonstrations}, Abu Dhabi, U.A.E.,
  November 2022. Association for Computational Linguistics.

\bibitem[\protect\citeauthoryear{Campos \bgroup \em et al.\egroup
  }{2017}]{campos2017pixels}
Victor Campos, Brendan Jou, and Xavier Giro-i Nieto.
\newblock From pixels to sentiment: Fine-tuning cnns for visual sentiment
  prediction.
\newblock {\em Image and Vision Computing}, 65:15--22, 2017.

\bibitem[\protect\citeauthoryear{Chauhan \bgroup \em et al.\egroup
  }{2019}]{chauhan2019context}
Dushyant~Singh Chauhan, Md~Shad Akhtar, Asif Ekbal, and Pushpak Bhattacharyya.
\newblock Context-aware interactive attention for multi-modal sentiment and
  emotion analysis.
\newblock In {\em Proceedings of the 2019 Conference on Empirical Methods in
  Natural Language Processing and the 9th International Joint Conference on
  Natural Language Processing (EMNLP-IJCNLP)}, pages 5647--5657, 2019.

\bibitem[\protect\citeauthoryear{Chen \bgroup \em et al.\egroup
  }{2014}]{chen2014deepsentibank}
Tao Chen, Damian Borth, Trevor Darrell, and Shih-Fu Chang.
\newblock Deepsentibank: Visual sentiment concept classification with deep
  convolutional neural networks.
\newblock {\em arXiv preprint arXiv:1410.8586}, 2014.

\bibitem[\protect\citeauthoryear{Diakopoulos and
  Shamma}{2010}]{diakopoulos2010characterizing}
Nicholas~A Diakopoulos and David~A Shamma.
\newblock Characterizing debate performance via aggregated twitter sentiment.
\newblock In {\em Proceedings of the SIGCHI conference on human factors in
  computing systems}, pages 1195--1198, 2010.

\bibitem[\protect\citeauthoryear{Dosovitskiy \bgroup \em et al.\egroup
  }{2020}]{dosovitskiy2020image}
Alexey Dosovitskiy, Lucas Beyer, Alexander Kolesnikov, Dirk Weissenborn,
  Xiaohua Zhai, Thomas Unterthiner, Mostafa Dehghani, Matthias Minderer, Georg
  Heigold, Sylvain Gelly, et~al.
\newblock An image is worth 16x16 words: Transformers for image recognition at
  scale.
\newblock {\em arXiv preprint arXiv:2010.11929}, 2020.

\bibitem[\protect\citeauthoryear{Gelli \bgroup \em et al.\egroup
  }{2015}]{gelli2015image}
Francesco Gelli, Tiberio Uricchio, Marco Bertini, Alberto Del~Bimbo, and
  Shih-Fu Chang.
\newblock Image popularity prediction in social media using sentiment and
  context features.
\newblock In {\em Proceedings of the 23rd ACM international conference on
  Multimedia}, pages 907--910, 2015.

\bibitem[\protect\citeauthoryear{Hazarika \bgroup \em et al.\egroup
  }{2020}]{hazarika2020misa}
Devamanyu Hazarika, Roger Zimmermann, and Soujanya Poria.
\newblock Misa: Modality-invariant and-specific representations for multimodal
  sentiment analysis.
\newblock In {\em Proceedings of the 28th ACM International Conference on
  Multimedia}, pages 1122--1131, 2020.

\bibitem[\protect\citeauthoryear{Islam and Zhang}{2016}]{islam2016visual}
Jyoti Islam and Yanqing Zhang.
\newblock Visual sentiment analysis for social images using transfer learning
  approach.
\newblock In {\em 2016 IEEE International Conferences on Big Data and Cloud
  Computing (BDCloud), Social Computing and Networking (SocialCom), Sustainable
  Computing and Communications (SustainCom)(BDCloud-SocialCom-SustainCom)},
  pages 124--130. IEEE, 2016.

\bibitem[\protect\citeauthoryear{Jansen \bgroup \em et al.\egroup
  }{2009}]{jansen2009twitter}
Bernard~J Jansen, Mimi Zhang, Kate Sobel, and Abdur Chowdury.
\newblock Twitter power: Tweets as electronic word of mouth.
\newblock {\em Journal of the American society for information science and
  technology}, 60(11):2169--2188, 2009.

\bibitem[\protect\citeauthoryear{Jou \bgroup \em et al.\egroup
  }{2015}]{jou2015visual}
Brendan Jou, Tao Chen, Nikolaos Pappas, Miriam Redi, Mercan Topkara, and
  Shih-Fu Chang.
\newblock Visual affect around the world: A large-scale multilingual visual
  sentiment ontology.
\newblock In {\em Proceedings of the 23rd ACM international conference on
  Multimedia}, pages 159--168, 2015.

\bibitem[\protect\citeauthoryear{Kaplan \bgroup \em et al.\egroup
  }{2020}]{kaplan2020scaling}
Jared Kaplan, Sam McCandlish, Tom Henighan, Tom~B Brown, Benjamin Chess, Rewon
  Child, Scott Gray, Alec Radford, Jeffrey Wu, and Dario Amodei.
\newblock Scaling laws for neural language models.
\newblock {\em arXiv preprint arXiv:2001.08361}, 2020.

\bibitem[\protect\citeauthoryear{Katsurai and Satoh}{2016}]{katsurai2016image}
Marie Katsurai and Shin'ichi Satoh.
\newblock Image sentiment analysis using latent correlations among visual,
  textual, and sentiment views.
\newblock In {\em 2016 IEEE International Conference on Acoustics, Speech and
  Signal Processing (ICASSP)}, pages 2837--2841. IEEE, 2016.

\bibitem[\protect\citeauthoryear{Khan and Fu}{2021}]{khan2021exploiting}
Zaid Khan and Yun Fu.
\newblock Exploiting bert for multimodal target sentiment classification
  through input space translation.
\newblock In {\em Proceedings of the 29th ACM International Conference on
  Multimedia}, pages 3034--3042, 2021.

\bibitem[\protect\citeauthoryear{Khosla \bgroup \em et al.\egroup
  }{2014}]{khosla2014makes}
Aditya Khosla, Atish Das~Sarma, and Raffay Hamid.
\newblock What makes an image popular?
\newblock In {\em Proceedings of the 23rd international conference on World
  wide web}, pages 867--876, 2014.

\bibitem[\protect\citeauthoryear{Li \bgroup \em et al.\egroup
  }{2012}]{li2012scaring}
Bing Li, Songhe Feng, Weihua Xiong, and Weiming Hu.
\newblock Scaring or pleasing: exploit emotional impact of an image.
\newblock In {\em Proceedings of the 20th ACM international conference on
  Multimedia}, pages 1365--1366, 2012.

\bibitem[\protect\citeauthoryear{Li \bgroup \em et al.\egroup
  }{2018}]{li2018image}
Zuhe Li, Yangyu Fan, Weihua Liu, and Fengqin Wang.
\newblock Image sentiment prediction based on textual descriptions with
  adjective noun pairs.
\newblock {\em Multimedia Tools and Applications}, 77:1115--1132, 2018.

\bibitem[\protect\citeauthoryear{Liu and Zhang}{2012}]{liu2012survey}
Bing Liu and Lei Zhang.
\newblock A survey of opinion mining and sentiment analysis.
\newblock In {\em Mining text data}, pages 415--463. Springer, 2012.

\bibitem[\protect\citeauthoryear{Loureiro \bgroup \em et al.\egroup
  }{2022}]{loureiro2022timelms}
Daniel Loureiro, Francesco Barbieri, Leonardo Neves, Luis~Espinosa Anke, and
  Jose Camacho-Collados.
\newblock Timelms: Diachronic language models from twitter.
\newblock {\em arXiv preprint arXiv:2202.03829}, 2022.

\bibitem[\protect\citeauthoryear{Machajdik and
  Hanbury}{2010}]{machajdik2010affective}
Jana Machajdik and Allan Hanbury.
\newblock Affective image classification using features inspired by psychology
  and art theory.
\newblock In {\em Proceedings of the 18th ACM international conference on
  Multimedia}, pages 83--92, 2010.

\bibitem[\protect\citeauthoryear{Mahajan \bgroup \em et al.\egroup
  }{2018}]{mahajan2018exploring}
Dhruv Mahajan, Ross Girshick, Vignesh Ramanathan, Kaiming He, Manohar Paluri,
  Yixuan Li, Ashwin Bharambe, and Laurens Van Der~Maaten.
\newblock Exploring the limits of weakly supervised pretraining.
\newblock In {\em Proceedings of the European conference on computer vision
  (ECCV)}, pages 181--196, 2018.

\bibitem[\protect\citeauthoryear{Mejova \bgroup \em et al.\egroup
  }{2013}]{mejova2013gop}
Yelena Mejova, Padmini Srinivasan, and Bob Boynton.
\newblock Gop primary season on twitter: " popular" political sentiment in
  social media.
\newblock In {\em Proceedings of the sixth ACM international conference on Web
  search and data mining}, pages 517--526, 2013.

\bibitem[\protect\citeauthoryear{Mintz \bgroup \em et al.\egroup
  }{2009}]{mintz2009distant}
Mike Mintz, Steven Bills, Rion Snow, and Dan Jurafsky.
\newblock Distant supervision for relation extraction without labeled data.
\newblock In {\em Proceedings of the Joint Conference of the 47th Annual
  Meeting of the ACL and the 4th International Joint Conference on Natural
  Language Processing of the AFNLP}, pages 1003--1011, 2009.

\bibitem[\protect\citeauthoryear{Peng \bgroup \em et al.\egroup
  }{2015}]{peng2015mixed}
Kuan-Chuan Peng, Tsuhan Chen, Amir Sadovnik, and Andrew~C Gallagher.
\newblock A mixed bag of emotions: Model, predict, and transfer emotion
  distributions.
\newblock In {\em Proceedings of the IEEE conference on computer vision and
  pattern recognition}, pages 860--868, 2015.

\bibitem[\protect\citeauthoryear{Robert}{1980}]{robert1980emotion}
Plutchik Robert.
\newblock Emotion: a psychoevolutionary synthesis.
\newblock {\em New York7 Harper and Row}, 1980.

\bibitem[\protect\citeauthoryear{Rui \bgroup \em et al.\egroup
  }{2013}]{rui2013whose}
Huaxia Rui, Yizao Liu, and Andrew Whinston.
\newblock Whose and what chatter matters? the effect of tweets on movie sales.
\newblock {\em Decision support systems}, 55(4):863--870, 2013.

\bibitem[\protect\citeauthoryear{Salur and Aydin}{2020}]{salur2020novel}
Mehmet~Umut Salur and Ilhan Aydin.
\newblock A novel hybrid deep learning model for sentiment classification.
\newblock {\em IEEE Access}, 8:58080--58093, 2020.

\bibitem[\protect\citeauthoryear{Siersdorfer \bgroup \em et al.\egroup
  }{2010}]{siersdorfer2010analyzing}
Stefan Siersdorfer, Enrico Minack, Fan Deng, and Jonathon Hare.
\newblock Analyzing and predicting sentiment of images on the social web.
\newblock In {\em Proceedings of the 18th ACM international conference on
  Multimedia}, pages 715--718, 2010.

\bibitem[\protect\citeauthoryear{Sun \bgroup \em et al.\egroup
  }{2017}]{sun2017revisiting}
Chen Sun, Abhinav Shrivastava, Saurabh Singh, and Abhinav Gupta.
\newblock Revisiting unreasonable effectiveness of data in deep learning era.
\newblock In {\em Proceedings of the IEEE international conference on computer
  vision}, pages 843--852, 2017.

\bibitem[\protect\citeauthoryear{Totti \bgroup \em et al.\egroup
  }{2014}]{totti2014impact}
Luam~Catao Totti, Felipe~Almeida Costa, Sandra Avila, Eduardo Valle, Wagner
  Meira~Jr, and Virgilio Almeida.
\newblock The impact of visual attributes on online image diffusion.
\newblock In {\em Proceedings of the 2014 ACM conference on Web science}, pages
  42--51, 2014.

\bibitem[\protect\citeauthoryear{Truong and Lauw}{2019}]{truong2019vistanet}
Quoc-Tuan Truong and Hady~W Lauw.
\newblock Vistanet: Visual aspect attention network for multimodal sentiment
  analysis.
\newblock In {\em Proceedings of the AAAI conference on artificial
  intelligence}, volume~33, pages 305--312, 2019.

\bibitem[\protect\citeauthoryear{Truong and Lauw}{2023}]{truong2023concept}
Quoc-Tuan Truong and Hady~W Lauw.
\newblock Concept-oriented transformers for visual sentiment analysis.
\newblock In {\em Proceedings of the Sixteenth ACM International Conference on
  Web Search and Data Mining}, pages 1111--1119, 2023.

\bibitem[\protect\citeauthoryear{Vadicamo \bgroup \em et al.\egroup
  }{2017}]{2017_ICCV}
Lucia Vadicamo, Fabio Carrara, Andrea Cimino, Stefano Cresci, Felice
  Dell'Orletta, Fabrizio Falchi, and Maurizio Tesconi.
\newblock Cross-media learning for image sentiment analysis in the wild.
\newblock In {\em Proceedings of the IEEE International Conference on Computer
  Vision (ICCV) Workshops}, Oct 2017.

\bibitem[\protect\citeauthoryear{Wu \bgroup \em et al.\egroup
  }{2020}]{wu2020visual}
Lifang Wu, Mingchao Qi, Meng Jian, and Heng Zhang.
\newblock Visual sentiment analysis by combining global and local information.
\newblock {\em Neural Processing Letters}, 51(3):2063--2075, 2020.

\bibitem[\protect\citeauthoryear{Yang \bgroup \em et al.\egroup
  }{2018}]{yang2018visual}
Jufeng Yang, Dongyu She, Ming Sun, Ming-Ming Cheng, Paul~L Rosin, and Liang
  Wang.
\newblock Visual sentiment prediction based on automatic discovery of affective
  regions.
\newblock {\em IEEE Transactions on Multimedia}, 20(9):2513--2525, 2018.

\bibitem[\protect\citeauthoryear{You \bgroup \em et al.\egroup
  }{2015}]{you2015robust}
Quanzeng You, Jiebo Luo, Hailin Jin, and Jianchao Yang.
\newblock Robust image sentiment analysis using progressively trained and
  domain transferred deep networks.
\newblock In {\em Twenty-ninth AAAI conference on artificial intelligence},
  2015.

\bibitem[\protect\citeauthoryear{You \bgroup \em et al.\egroup
  }{2016}]{you2016building}
Quanzeng You, Jiebo Luo, Hailin Jin, and Jianchao Yang.
\newblock Building a large scale dataset for image emotion recognition: The
  fine print and the benchmark.
\newblock In {\em Proceedings of the AAAI conference on artificial
  intelligence}, volume~30, 2016.

\bibitem[\protect\citeauthoryear{Yu \bgroup \em et al.\egroup
  }{2021}]{yu2021learning}
Wenmeng Yu, Hua Xu, Ziqi Yuan, and Jiele Wu.
\newblock Learning modality-specific representations with self-supervised
  multi-task learning for multimodal sentiment analysis.
\newblock In {\em Proceedings of the AAAI conference on artificial
  intelligence}, volume~35, pages 10790--10797, 2021.

\end{thebibliography}
